\documentclass[fleqn,10pt]{wlscirep}
\usepackage[utf8]{inputenc}
\usepackage[T1]{fontenc}
\usepackage{svg}
\usepackage{algorithm}
\usepackage{algpseudocode}
\usepackage{float}
\usepackage{amsmath}   

\usepackage[justification=justified,singlelinecheck=false]{caption}
\usepackage{comment}
\usepackage{xcolor}
\usepackage{multirow}

\title{CRISP -- Clustering-Based Redundancy-Reduced Instance Sampling for Pathology Case Representation and Retrieval}

\author[1,2]{Zahra Rahimi Afzal}
\author[1]{Wataru Uegami, MD, PhD}
\author[1]{Saghir Alfasly, PhD}
\author[3]{Wenchao Han, PhD}
\author[4]{Saba Yasir, MD.}
\author[5]{Judy C. Boughey, MD}
\author[6]{Matthew P. Goetz, MD}
\author[7]{Krishna R. Kalari, PhD}
\author[1,4,*]{H.R. Tizhoosh, PhD}
\affil[1]{Kimia Lab, Department of Artificial Intelligence \& Informatics, Mayo Clinic, Rochester, MN, USA }
\affil[2]{DICE Lab, Department of Electrical and Computer Engineering, University of Illinois Chicago, IL, USA }
\affil[3]{Division of Computational Pathology and Informatics, Mayo Clinic, Rochester, MN, USA}
\affil[4]{Department of Laboratory Medicine and Pathology, Mayo Clinic, Rochester, MN, USA}
\affil[5]{Department of Breast and Melanoma Surgical Oncology, Comprehensive Cancer Center, Mayo Clinic, Rochester, MN, USA}
\affil[6]{Department of Oncology, Comprehensive Cancer Center, Mayo Clinic, Rochester, MN, USA}
\affil[76]{Department of Quantitative Health Sciences, Mayo Clinic, Rochester, MN, USA}
\affil[*]{Corresponding author: tizhoosh.hamid@mayo.edu}

\begin{abstract}
Digital pathology archives increasingly contain multiple whole-slide images (WSIs) per case, capturing spatially distinct tumor regions and reflecting intrinsic morphological heterogeneity. However, most existing approaches rely on a single pathologist-selected slide, thereby discarding potentially informative evidence distributed across the remaining WSIs. To date, no autonomous framework has been proposed for comprehensive multi-WSI case processing. Here, we present an unsupervised framework for case-level analysis that integrates information from all available slides within a case. Rather than relying on a single designated slide, the proposed approach constructs case-level representations by selectively distilling informative patches across WSIs. We introduce Clustering-Based Redundancy-Reduced Instance Sampling for Pathology (CRISP), a two-stage framework that first reduces redundancy within individual WSIs and subsequently applies clustering-based sampling to select a compact yet representative set of patches for the entire case. The resulting patch set captures case-level heterogeneity while avoiding exhaustive processing of gigapixel images, and directly serves as a retrieval index. Using two Mayo Clinic breast cancer datasets for diagnosis and treatment planning, we demonstrate that CRISP consistently matches or surpasses the current standard practice of combined model and pathologist slide selection for patient/case search and retrieval. By automating case-level processing and eliminating subjective WSI selection, CRISP potentially enables the exploitation of clinically relevant information distributed across multiple WSIs that is currently overlooked.
\end{abstract}

\begin{document}

\flushbottom
\maketitle

\section*{Main}
Histopathology archives contain a rich record of morphological variation that can inform diagnosis, rare disease recognition, treatment planning, and retrospective analysis. Content-based image retrieval offers a direct mechanism for leveraging these repositories by matching newly acquired specimens to previously diagnosed cases with similar morphological patterns~\cite{madabhushi2016digital, kalra2020yottixel}. Such systems have the potential to support clinical decision-making, reduce diagnostic uncertainty, and facilitate large-scale data-driven pathology~\cite{tizhoosh2024image, tizhoosh2021searching}. However, despite substantial advances in computational pathology, retrieval at scale remains constrained by a fundamental mismatch between the size and heterogeneity of whole-slide images~(WSIs) and the need for compact, comparable representations~\cite{lahr2024analysis}.

A central challenge is therefore not only feature extraction, but representation: which regions of a WSI should define a case? Early retrieval systems partitioned slides into tiles and performed matching using engineered or learned descriptors~\cite{barker2016automated,zhang2015hashing,hegde2019smily}.
Subsequent weakly supervised and multiple-instance learning approaches enabled slide-level modeling, with attention-based pooling methods such as ABMIL~\cite{abmil} and transformer-based architectures including CLAM~\cite{lu2021clam} and TransMIL~\cite{shao2021transmil} identifying diagnostically relevant regions without dense annotation~\cite{campanella2019clinical}. ABMIL~\cite{abmil} learns a gated attention pooling mechanism over patch features to generate slide-level representations from weakly labelled bags. 
More recently, foundation models pretrained on large-scale pathology corpora have substantially improved feature quality and transferability across downstream tasks~\cite{chen2024uni,lu2024conch,vorontsov2024virchow,xu2024gigapath,wang2024chief,zimmermann2024virchow2,ding2025titan}.  
Extending beyond slide-centric representations, MOOZY~\cite{moozy} introduced a patient-level model that encodes WSIs collections into unified case-level representations. However, such approaches rely on large-scale pretraining and substantial computational resources, and do not directly address how multiple WSIs can be efficiently consolidated into compact and interpretable representations for patient retrieval. Consequently, most existing retrieval frameworks remain fundamentally slide-centric, relying on a single pathologist-selected WSI as a surrogate for the patient case.

Most existing retrieval frameworks therefore compress \emph{one WSI} into a representative patch set. Yottixel, for example, constructs compact slide mosaics using hierarchical clustering~\cite{kalra2020yottixel}, whereas SPLICE sequentially suppresses redundant regions to retain morphologically diverse tissue patterns~\cite{alsaafin2024splice}. These approaches substantially reduce computational burden and storage requirements while maintaining retrieval performance. However, they implicitly treat an individual slide as a complete proxy for the case and therefore ignore the broader spatial and morphological context distributed across multiple WSIs.

This assumption is inconsistent with routine clinical workflow. In diagnostic pathology, patient cases are commonly represented by multiple slides originating from different tissue blocks, tumor regions, staining preparations, or sampling depths. These slides may capture complementary information, including heterogeneous tumor architectures, invasive fronts, in situ components, stromal responses, necrosis, and coexisting lesions. Consequently, selecting a single slide can discard informative evidence and introduces observer-dependent variability associated with manual slide selection. This limitation is particularly pronounced in breast pathology, where heterogeneous growth patterns and mixed histological subtypes can span multiple sections and may not be adequately summarized by a single WSI.

To address this limitation, we introduce a \textbf{case-level retrieval} framework that processes all available WSIs to construct a comprehensive case representation. We propose Clustering-Based Redundancy-Reduced Instance Sampling for Pathology (CRISP), a two-stage framework that first applies SPLICE to reduce redundancy within each WSI and subsequently applies $k$-means clustering across slides to select a compact set of patches representing the entire case. Rather than exhaustively processing gigapixel images, CRISP selectively distills informative and diverse tissue regions into a concise case-level descriptor that directly serves as a retrieval index and can additionally support downstream tasks such as classification and outcome prediction. By consolidating information across slides, the framework captures case-level heterogeneity while maintaining computational efficiency.

We evaluate CRISP on two Mayo Clinic breast cancer cohorts spanning diagnostic retrieval and treatment-planning applications. Compared with conventional single-slide representations, the proposed framework matches or surpasses existing model-plus-pathologist workflows, in which a pathologist selects the WSI used by the model. Beyond performance gains, our results demonstrate that case-level modeling provides a practical and scalable alternative to slide-centric indexing for heterogeneous diseases. More broadly, CRISP establishes a foundation for autonomous multi-WSI computational pathology pipelines that leverage the full morphological context of a patient case rather than a manually selected subset of slides.

\section{Results}

\begin{figure}[htbp]
    \centering
    \includegraphics[width=0.80\linewidth]{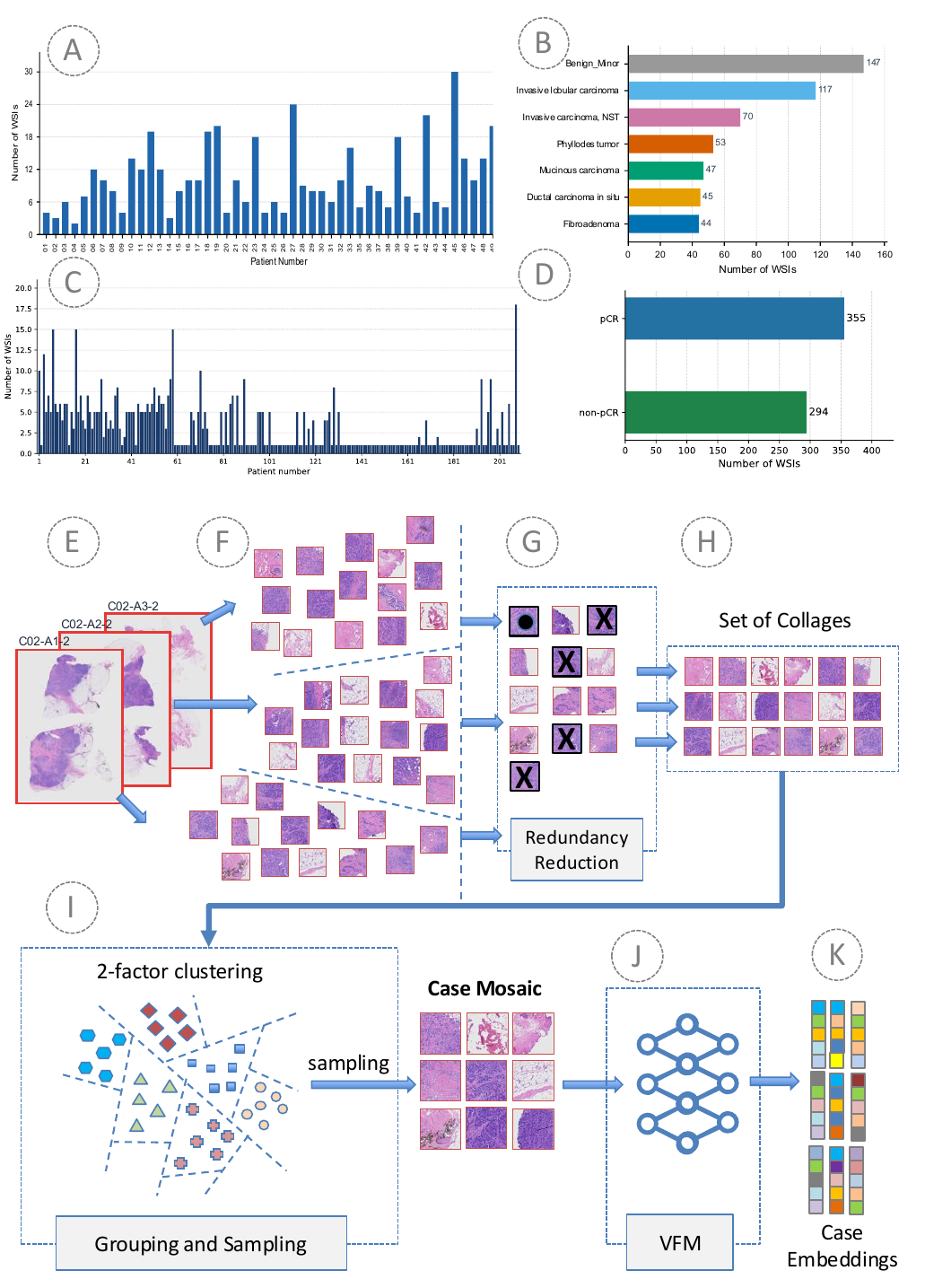}
    \caption{CRISP workflow for case-level histopathology representation from multi-slide whole-slide images:
(A) Number of whole-slide images (WSIs) per case in the \textbf{diagnostic study cohort}. (B) Number of WSIs per diagnostic subtype, showing class imbalance across categories. (C) Number of whole-slide images (WSIs) per triple-negative breast cancer patient in the \textbf{treatment study cohort}. (D) Number of patients for positive and negative pathologic complete response (pCR). (E) Example patient (C02) with three WSIs used as case input. (F) Initial candidate patches sampled from tissue-rich regions of each WSI. (G) SPLICE-based pruning removes redundant or low-information patches (crossed tiles), retaining diverse, informative patches. (H) Aggregation of retained patches into a reduced case-level patch set across all WSIs from the same case. (I) Construction of Yottixel’s case mosaic via two-stage clustering and representative sampling. (J) Extraction of patch features using a vision foundation model (VFM). (K) Assembly of the final case embedding from patch-level features for downstream retrieval and classification.}
    \label{fig:crisp}
\end{figure}


Fig.~\ref{fig:crisp} illustrates the two-stage CRISP pipeline and the datasets 
used for evaluation. Two cohorts were included: a diagnostic breast cancer dataset 
(50~patients, 523~WSIs, 7~subtypes, mean ${\approx}10.5$~WSIs per case) and 
a triple negative breast cancer (TNBC) treatment cohort (209~patients, 649~WSIs, mean ${\approx}3.1$~WSIs per case), in which 120~patients (57.4\%) 
achieved pathological complete response (pCR) following neoadjuvant chemotherapy and 
89~(42.6\%) did not. All experiments used leave-one-patient-out cross-validation 
with  embeddings of a foundation model (Virchow2 CLS)~\cite{zimmermann2024virchow2} and cosine and Euclidean patinet matching. Full details are given in Methods.

\subsection*{case-level retrieval improves performance for both diagnosis and treatment response prediction}
\begin{figure}[htb]
    \centering
    \includegraphics[width=0.49\linewidth]{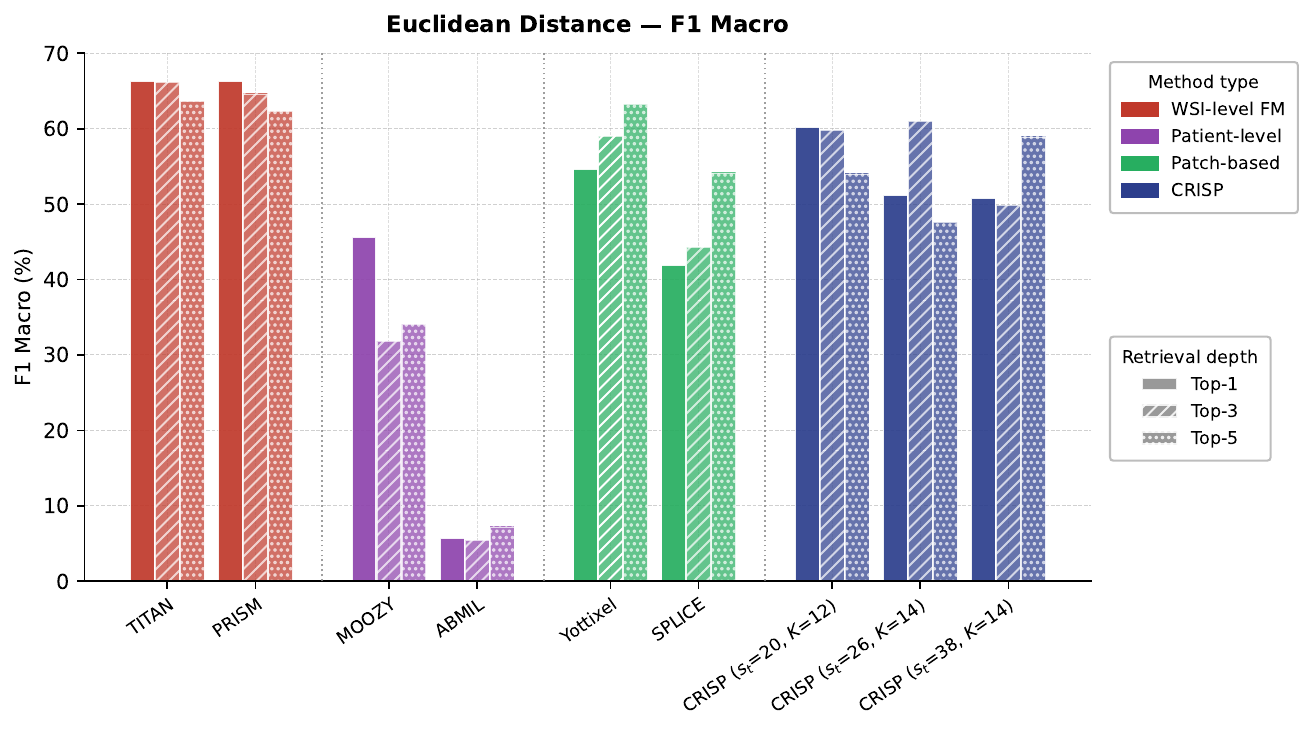}
    \includegraphics[width=0.49\linewidth]{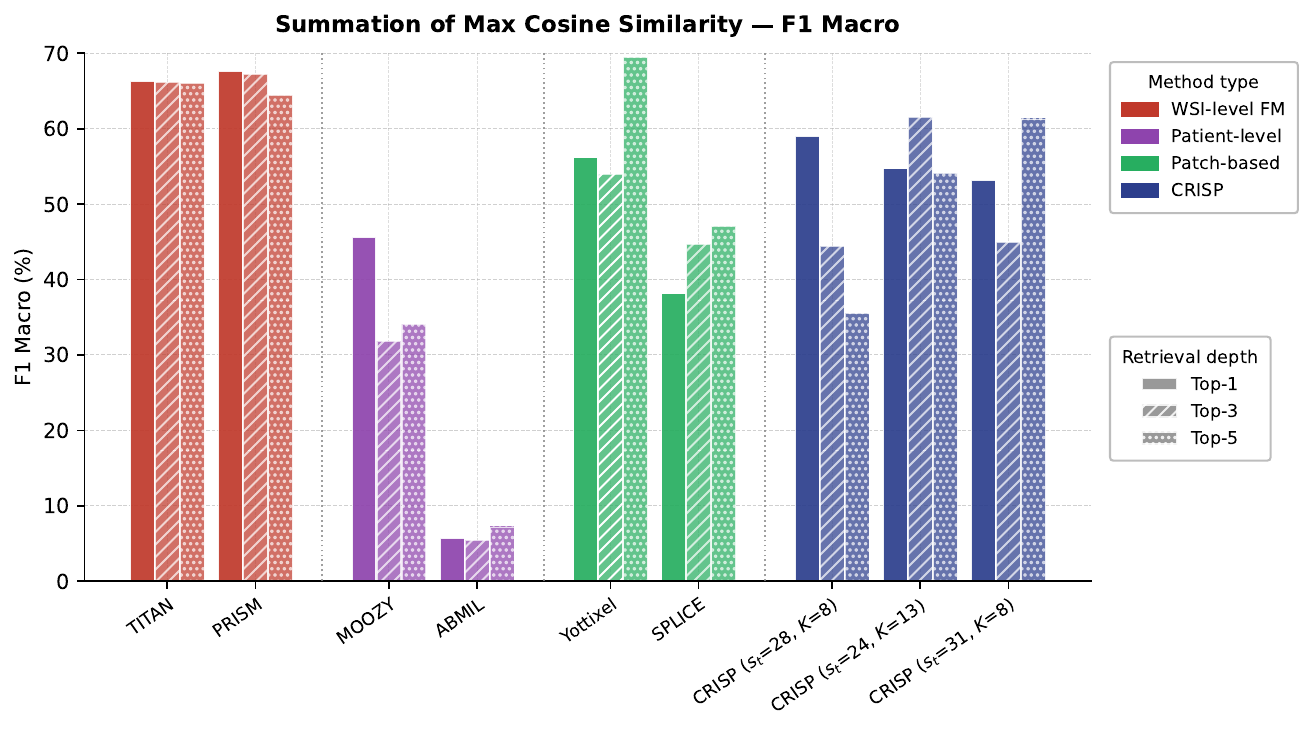}
    \caption{Results for breast morphological subtyping (50 patients):  
    Yottixel, SPLICE, and two WSI foundation models TITAN and PRISM were applied on a single WSI selected by a pathologist as baselines; MOOZY and ABMIL are included as patient-level representation methods that aggregate all case WSIs into a single embedding without explicit patch diversity; compared to CRISP in different settings to process all case WSIs automatically. 
    Macro-averaged F1 (\%) at Top-1, Top-3, and Top-5. Bold denotes the best value per row.}
    \label{fig:diagtest}
\end{figure}

Figure~\ref{fig:diagtest} reports macro-averaged F1 for all 
configurations for a dataset of 50 patients. At Top-1, the configuration $s_t{=}20$, $K{=}12$, 
$\alpha{=}3.5\%$ reached 60.2\%, against 54.6\% for 
Yottixel$+$Pathologist~(\textbf{+5.6~pp}) and 41.9\% for SPLICE$+$Pathologist~(\textbf{+18.3~pp}). 
At Top-3, the same configuration achieved 59.8\%, outperforming
Yottixel$+$Pathologist (59.0\%; +0.8~pp) and
SPLICE$+$Pathologist (44.3\%; +15.5~pp).

At Top-3, the configuration $s_t{=}26$, $K{=}14$, $\alpha{=}9.0\%$ 
reached 61.0\%, against 59.0\% for Yottixel$+$Pathologist~(+2.0~pp) and 
44.3\% for SPLICE$+$Pathologist~(+16.7~pp). 
The optimal hyperparameters differed by ranking depth: at Top-1, 
lower SPLICE thresholds retain more patches per slide, concentrating 
the representation around precise nearest-neighbour matches; 
at Top-3, higher thresholds with more clusters broaden the 
morphological coverage, supporting majority agreement. At Top-5, 
no case-level configuration reached Yottixel$+$Pathologist~(63.2\%); 
the best was 58.9\%, a gap that narrows as neighbourhood size 
grows but does not close.

Every case-level configuration surpassed SPLICE$+$Pathologist at all 
ranking depths, regardless of hyperparameter choice. case-level 
consolidation is most effective at Top-1 and Top-3; at Top-5, 
curated single-slide specificity retains a partial advantage.
The grid search spanned $s_t \in \{20,\ldots,40\}$, $K \in 
\{7,\ldots,20\}$, and $\alpha \in [0.25\%, 10\%]$. High-performing configurations clustered 
around $s_t{=}20$--26 and $K{=}12$--14 across a wide range of 
$\alpha$ values, with no single combination standing out as 
uniquely optimal.

We compared our cosine-similarity retrieval approach against two baselines: Yottixel+Pathologist and CRISP evaluated under median-of-minimus processing. Using sum-of-maximum cosine similarities as the retrieval score, our method achieves a macro-F1 of 59.0\% at Top-1, 61.5\% at Top-3, and 61.3\% at Top-5. Compared to Yottixel+Pathologist, cosine retrieval improves Top-1 by 4.4 percentage points and Top-3 by 1.5 percentage points, while narrowing the Top-5 gap to just -1.9 points. Compared to the median-based CRISP baseline ($s_t{=}38$, $K{=}14$, $\alpha{=}8.3\%$), cosine aggregation yields consistent gains across all retrieval depths (+8.2, +10.7, and +2.4 percentage points at Top-1, Top-3, and Top-5,  respectively), demonstrating that summing per-patch maximum cosine affinities provides a more discriminative case-level retrieval score than the median of minimum distances.

TITAN and PRISM, WSI-level foundation models trained on tens of thousands of whole-slide images, serve as upper-bound references, consistently scoring above 65\% under Euclidean distance at Top-1 (Figure~\ref{fig:diagtest}, left). Both models, however, depend on a pathologist-curated single representative slide per case, a requirement that constrains their deployment in fully automated settings and makes them subject to inter- and intra-observer variability. The tumor heterogeneity is not noise to be smoothed away; in many breast morphological subtypes, it is diagnostically decisive, and aggregation-based encoders structurally cannot preserve it. Compressing the full spatial extent of a WSI, or worse WSIs, into a single fixed-length embedding forces the model to collapse morphologically distinct tissue regions into a single averaged descriptor, discarding the very focal signals that differentiate subtypes. This compression penalizes retrieval results: Yottixel+Pathologist, which forgoes global aggregation in favor of a sparse mosaic of locally representative patches, surpasses both TITAN and PRISM under summation of maximum cosine similarity at Top-5, reaching approximately 70\% against their 65--67\% range (Figure~\ref{fig:diagtest}, right). Our approach carries neither of these constraints; it requires no expert slide selection and operates directly on patch-level affinities across all available WSIs per case. That our best configurations are close to within a few percentage points of TITAN and PRISM across most retrieval depths, therefore, speaks to the practical value of case-level consolidation, achieved without large-scale foundation model pre-training and without any pathologist involvement.
Interobserver variability in breast pathology is well documented, with $\kappa$ statistics as low as 0.34 for proliferative intraductal lesions among experienced pathologists, a level conventionally classified as only fair agreement~\cite{jain2011adh}, and broader analyses confirm that such variability persists even among specialist breast pathologists across a range of challenging diagnostic categories~\cite{allison2014variability}. No published literature, however, addresses the variability introduced when a pathologist selects a single representative WSI for AI training or retrieval, leaving the reproducibility of that curation step an uncharacterized and largely unacknowledged source of noise in methods that depend upon it.

Patient-level representation learning alone was insufficient to achieve competitive retrieval performance. MOOZY, despite being pretrained as a patient-level  model, achieved substantially lower F1 scores than both pathologist-guided baselines and CRISP across all retrieval settings. Similarly, ABMIL, which is trained from scratch within each leave-one-patient-out fold, performed poorly, likely owing to the limited cohort size and the difficulty of learning robust bag-level representations from weak supervision. In contrast, CRISP consistently outperformed both MOOZY and ABMIL, demonstrating that explicit redundancy reduction and diversity-preserving patch selection across all available WSIs provides a more effective representation for patient retrieval.

\begin{figure}[htb]
    \centering
    \includegraphics[width=0.49\linewidth]{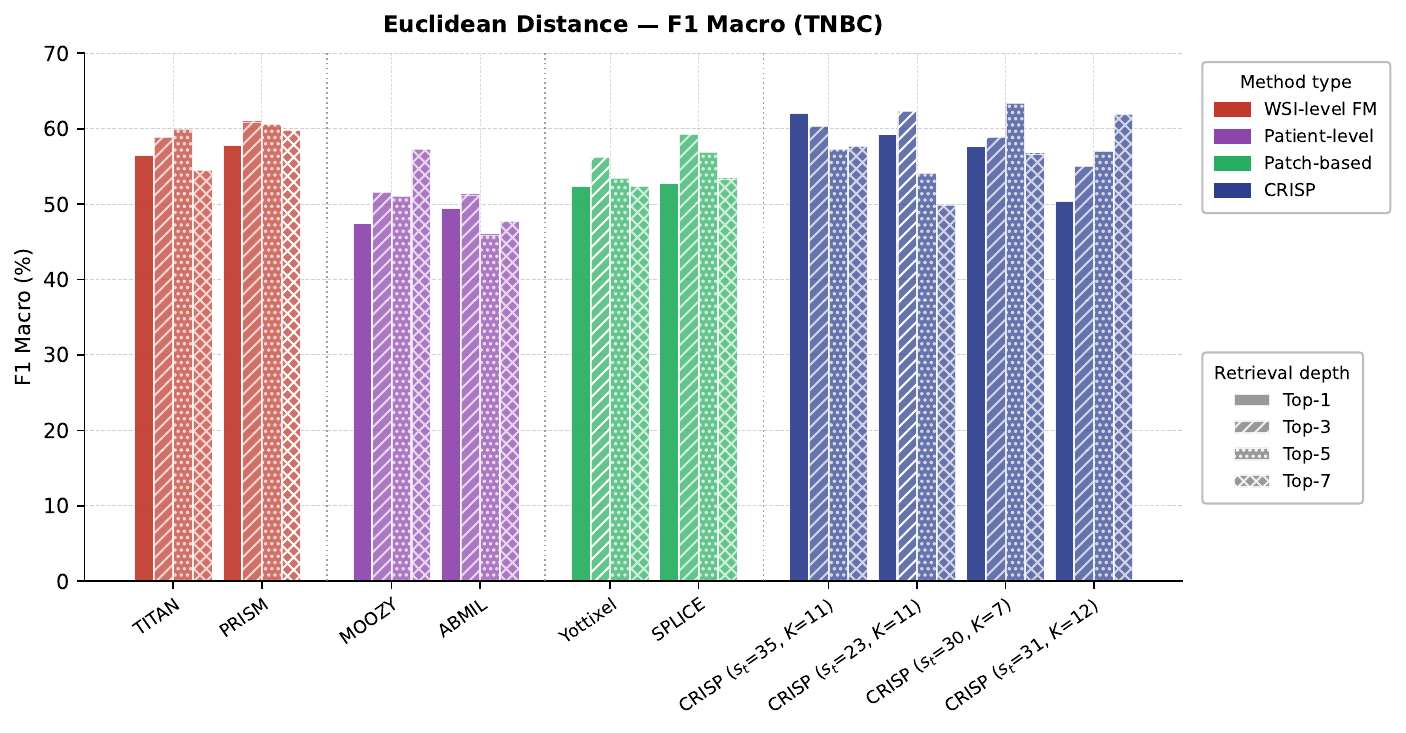}
    \includegraphics[width=0.49\linewidth]{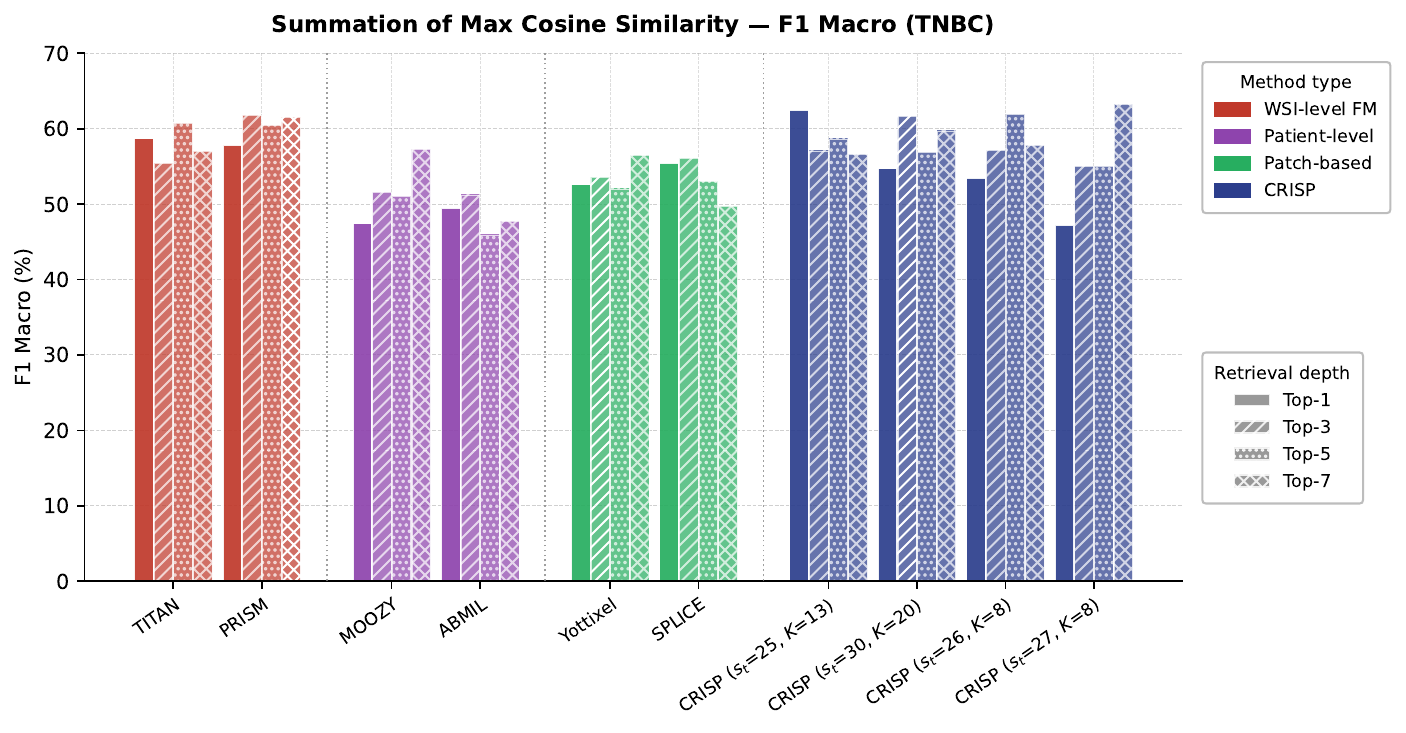}
    \caption{Results for prediction of treatment response for triple negative breast cancer (209 patients): 
    Yottixel, SPLICE, and two WSI foundation models TITAN and PRISM were applied on a single WSI selected by a pathologist as baselines; MOOZY and ABMIL are included as patient-level representation methods that aggregate all case WSIs into a single embedding without explicit patch diversity; compared to CRISP in different settings to process all case WSIs automatically. Macro-averaged F1 (\%) for leave-one-patient-out retrieval at Top-$N$.     Bold indicates the best value per row.}
    \label{fig:tnbctest}
\end{figure}

Figure~\ref{fig:tnbctest} presents the retrieval results in our TNBC cohort 
in two baseline categories: SPLICE$+$Pathologist, a morphology-driven 
retrieval system with pathologist-guided feature curation, and two large-scale 
WSI-level foundation models, Titan \cite{ding2025titan} and PRISM \cite{shaikovski2024prism}, pretrained on large corpora of whole-slide images. Performance is reported as macro-averaged F1 
under leave-one-patient-out evaluation at four retrieval depths 
($N \in \{1, 3, 5, 7\}$).
At Top-1, the most challenging operating point for clinical decision support, 
CRISP ($s_t{=}25$, $K{=}13$, $\alpha{=}0.25\%$) reached a macro-F1 of 62.5\%, 
outperforming Titan (58.7\%) and PRISM (57.8\%) by \textbf{3.8} and \textbf{4.7}~pp, respectively. 
These margins are meaningful in the context of binary pCR prediction, where a 
single incorrect retrieval directly affects the top-ranked candidate label. 
At Top-3, CRISP achieved 61.6\%, closely tracking PRISM (61.8\%) while 
exceeding Titan by \textbf{6.2}~pp. Performance advantages over PRISM persisted at 
greater retrieval depths ($+$\textbf{1.5}~pp at Top-5, $+$\textbf{1.8}~pp at Top-7), and the 
lead over Titan widened to \textbf{1.3} and \textbf{6.3}~pp at those same points. Relative 
to SPLICE$+$Pathologist, improvements reached 13.5~pp at Top-7, reflecting 
the greater discriminative capacity of deep embeddings over handcrafted 
descriptors in a heterogeneous, multi-slide cohort.

Across configurations, no single set of CRISP hyperparameters dominated 
uniformly. Tighter SPLICE thresholds with fewer $k$-means clusters concentrate 
selection on the most discriminative patches, favoring precise Top-1 retrieval; 
broader configurations distribute coverage across the slide and improve recall 
at larger $N$. This behavior is an interpretable property of the patch-selection 
design rather than instability: the optimal configuration depends on whether the 
intended workflow prioritizes a single high-confidence candidate or a ranked 
shortlist for pathologist review. Across all retrieval depths, at least one 
CRISP variant matched or exceeded both foundation models, establishing that 
a principled mosaic strategy, without any large-scale external pretraining,
is competitive on this diagnostically demanding TNBC cohort.

\subsection*{Diversity-aware clustering is critical at the case level}

To isolate the contribution of the case-level consolidation 
mechanism, we compared two reduction strategies on identical SPLICE 
 pools: sequential SPLICE re-selection over the pooled case 
patches and $k$-means clustering. $k$-means consistently 
outperformed SPLICE re-selection across all matched settings. 
case-level SPLICE re-selection remained in the 37--42\% Top-1 
macro-F1 range, while $k$-means reached 51--60\% at the same 
slide-level configurations. At $s_t{=}20$, the switch from
SPLICE to  $k$-means alone gained approximately 
20~percentage points in Top-1 macro-F1.

Once slide-level redundancy has been suppressed by CRISP, a second
sequential pruning pass at the case level is ineffective.
Explicitly partitioning the pooled patches into color-similar
clusters preserve inter-slide morphological diversity rather than
collapsing the representation onto a visually redundant subset.

\subsection*{CRISP case-level representations preserve diagnostic information}
Table~\ref{tab:patch_reduction_50pt} reports patch compression across the
50-patient multi-class histopathology cohort spanning seven disease categories.
TITAN and PRISM, operating at the WSI level, forward all tissue-filtered
patches without selection, retaining the full $235.0$ patches per WSI on
average and achieving no reduction.
Yottixel and SPLICE each work from a single key-slide WSI per case,
distilling $235.0$ raw patches down to roughly 14 and 12 representative
patches, respectively, a compression of ${\sim}94$--$95\%$.
CRISP draws from the complete multi-slide record: on average $2{,}393$
tissue-filtered patches per case, pooled across 509~WSIs
(119{,}634 patches in total across the 50-patient cohort).
A two-stage pipeline, SPLICE independently on each WSI, followed by $k$means across all WSIs of the same case, then distills this richer
input into a compact case representation.
The best CRISP configuration ($s_t{=}31$, $K{=}8$, $\alpha{=}3.5\%$)
retains only $8.2$ patches per case, a \textbf{99.7\%} reduction
relative to the full raw pool.

Table~\ref{tab:patch_reduction_data2} illustrates the computational compression
achieved by each method on the 209-patient TNBC cohort.
For the WSI-level foundation models TITAN and PRISM, no patch-level selection
is performed and all tissue patches are forwarded to the retrieval stage,
resulting in 0\% reduction.
Yottixel and SPLICE operate on single key-slide WSIs, each containing on average
$171.2$ raw tissue-filtered patches per case, and reduce this to approximately
10--11 representative patches through $k$means and SPLICE selection
respectively, corresponding to a ${\sim}94\%$ reduction.
CRISP, by contrast, has access to the full multi-WSI archive: on average
$525.0$ raw patches per case pooled across all available slides
(109{,}725 patches in total across 641~WSIs).
The best CRISP configuration ($s_t{=}30$, $K{=}7$, $\alpha{=}8.5\%$) retains
only $7.1$ patches per case, achieving a $\mathbf{98.6\%}$ reduction.
These results demonstrate that CRISP not only exploits multi-slide context
unavailable to single-WSI baselines, but does so with greater compression,
yielding a case mosaic that is simultaneously compact, computationally
efficient, and retrieval-effective.
The exact numerical values shown in Figures~\ref{fig:diagtest} and~\ref{fig:tnbctest} are provided in Tables~\ref{tab:overall_data1} and~\ref{tab:overall_tnbc}, respectively.

\begin{table}[t]
\centering
\begin{tabular}{p{4.9cm}ccc}
\toprule
Method & Top-1 F1 & Top-3 F1 & Top-5 F1 \\
\midrule
\multicolumn{4}{c}{Euclidean distance}\\ 
\midrule
\midrule
TITAN$+$Pathologist  & \textbf{66.30} & 66.17 & 63.60 \\
PRISM$+$Pathologist  & \textbf{66.28} & 64.61 & 62.25 \\
\midrule
MOOZY & \textbf{45.59} & 31.84 & 34.03\\
ABMIL  &  5.71          &  5.39          &  7.21 \\
\midrule
Yottixel$+$Pathologist & 54.63 & 58.99 & \textbf{63.23} \\
SPLICE$+$Pathologist  & 41.94 & 44.29 & \textbf{54.13} \\
\midrule
CRISP ($s_t{=}20$, $K{=}12$, $\alpha{=}3.5\%$)
    & \textbf{60.2} & 59.8 & 54.0 \\
CRISP ($s_t{=}26$, $K{=}14$, $\alpha{=}9.0\%$)
    & 51.2 & \textbf{61.0} & 47.6 \\
CRISP ($s_t{=}38$, $K{=}14$, $\alpha{=}8.3\%$)
    & 50.8 & 49.8 & \textbf{58.9} \\
\midrule
\midrule
\multicolumn{4}{c}{Summation max cosine similarity}  \\
\midrule
\midrule
TITAN$+$Pathologist  & \textbf{66.30} & 66.17 & 66.01 \\
PRISM$+$Pathologist  & \textbf{67.67} & 67.17 & 64.44 \\
\midrule
MOOZY & \textbf{45.59} & 31.84 & 34.03\\
ABMIL  &  5.71          &  5.39          &  7.21 \\
\midrule
Yottixel$+$Pathologist & 56.15 & 53.89 & \textbf{69.45} \\
SPLICE$+$Pathologist &38.12& 44.61 & \textbf{47.03}\\
\midrule
CRISP ($s_t{=}28$, $K{=}8$, $\alpha{=}9.75\%$)
    & \textbf{59.0} & 44.4 & 35.5 \\
CRISP ($s_t{=}24$, $K{=}13$, $\alpha{=}5.75\%$)
    & 54.7 & \textbf{61.5} & 54.1 \\
CRISP ($s_t{=}31$, $K{=}8$,\ \ $\alpha{=}3.5\%$)
    & 53.2 & 44.9 & \textbf{61.3} \\
\bottomrule
\end{tabular}
\caption{Results for breast morphological subtyping: Yottixel, SPLICE, and two WSI foundation models TITAN and PRISM applied on a single WSI selected by a pathologist as baseline compared to CRISP in different settings to process all case WSIs automatically; macro-averaged F1 (\%) at Top-1, Top-3, and Top-5. 
We additionally compare against the patient-level foundation model MOOZY and the supervised multiple-instance learning method ABMIL. CRISP consistently outperforms both MOOZY and ABMIL.
Bold denotes the best value per row.}
\label{tab:overall_data1}
\end{table}

\begin{table}[t]
\centering
\begin{tabular}{p{4.9cm}cccc}
\toprule
Method & Top-1 F1 & Top-3 F1 & Top-5 F1 & Top-7 F1\\
\midrule
\multicolumn{5}{c}{Euclidean distance}\\ 
\midrule
\midrule
TITAN$+$Pathologist  & 56.46 & 58.89 & \textbf{59.85} & 54.51 \\
PRISM$+$Pathologist  & 57.80 & \textbf{60.91} & 60.61 & 59.77 \\
\midrule
MOOZY  & 47.41 & \textbf{51.59} & 51.04 & 57.26 \\
ABMIL  & 49.42          & \textbf{51.23}          & 45.91          & 47.72 \\
\midrule
Yottixel$+$Pathologist & 52.41 & \textbf{56.18} & 53.44 & 52.33\\
SPLICE$+$Pathologist  & 52.82 & \textbf{59.21} & 56.82 & 53.34 \\
\midrule
CRISP ($s_t{=}35$, $K{=}11$, $\alpha{=}9.25\%$)
    & \textbf{62.03} & 60.28 & 57.22 & 57.61 \\   
CRISP ($s_t{=}23$, $K{=}11$, $\alpha{=}0.25 \%$)
    & 59.21 & \textbf{62.32} & 54.12 & 49.87 \\  
CRISP ($s_t{=}30$, $K{=}7$,\ \ $\alpha{=}8.50 \%$)
    & 57.62 & 58.87 & \textbf{63.33}& 56.66  \\
CRISP ($s_t{=}31$, $K{=}12$, $\alpha{=}9.75 \%$)
    & 50.36  & 55.05 & 56.97 & \textbf{61.93} \\
\midrule
\midrule
\multicolumn{5}{c}{Summation max cosine similarity}\\ 
\midrule
\midrule
TITAN$+$Pathologist  & 58.68 & 55.40 & \textbf{60.65} & 56.95 \\
PRISM$+$Pathologist  & 57.80 & \textbf{61.76} & 60.45 & 61.45 \\
\midrule
MOOZY  & 47.41 & \textbf{51.59} & 51.04 & 57.26 \\
ABMIL  & 49.42          & \textbf{51.23}          & 45.91          & 47.72 \\
\midrule
Yottixel$+$Pathologist & 52.57  & 53.59 & 52.01 & \textbf{56.46} \\
SPLICE$+$Pathologist  & 55.41 & \textbf{56.03} & 52.96 & 49.72 \\
\midrule
CRISP ($s_t{=}25$, $K{=}13$, $\alpha{=}0.25\%$)
    & \textbf{62.5} & 57.05 & 58.65 & 56.64 \\
CRISP ($s_t{=}30$, $K{=}20$, $\alpha{=}0.25\%$)
    & 54.7 & \textbf{61.6} & 56.9 & 59.7 \\
CRISP ($s_t{=}26$, $K{=}8$,\ \ $\alpha{=}9.75\%$)
    & 53.4 & 57.1 & \textbf{61.9}& 57.8 \\
CRISP ($s_t{=}27$, $K{=}8$, $\alpha{=}4.5\%$)
    & 47.2 & 55.00 & 55.02 & \textbf{63.2} \\
\bottomrule
\end{tabular}
\caption{Results for prediction of treatment response for triple negative breast cancer (209 patients): Yottixel, SPLICE, and two WSI foundation models TITAN and PRISM applied on a single WSI selected by a pathologist as baseline compared to CRISP in different settings to process all case WSIs automatically; macro-averaged F1 (\%) for leave-one-patient-out retrieval at Top-$N$. 
We additionally compare against the patient-level foundation model MOOZY and the supervised multiple-instance learning method ABMIL. 
Bold indicates the best value per row.}
\label{tab:overall_tnbc}
\end{table}

\begin{table}[t]
\centering
\begin{tabular}{lccc}
\toprule
Setting
    & $\bar{N}^{\text{Total}}$
    & $\bar{N}^{\text{Reduction}}$
    & Reduction \\
\midrule
Yottixel+Pathologist
    & 235  & 14  & 94.0\% \\
SPLICE+Pathologist
    & 235  & 12  & 94.8\% \\
TITAN+Pathologist
    & 235  & 235 & 0.0\%  \\
PRISM+Pathologist
    & 235  & 235 & 0.0\%  \\
\midrule
CRISP ($s_t{=}31,\,K{=}8,\,\alpha{=}3.5\%$)
    & 2{,}393 & 8  & \textbf{99.7\%} \\
\bottomrule
\end{tabular}
\caption{%
    Patch reduction statistics for the 50-patient histopathology cohort
    (7 disease classes).
    $\bar{N}^{\text{Total}}$ denotes the average number of raw tissue-filtered
    patches per case before any selection is applied;
    CRISP pools all available multi-WSI data per case
    (119{,}634 raw patches across 509~WSIs, $\bar{N}^{\text{Total}}=2{,}393$),
    whereas baselines operate on a single key-slide WSI per case
    ($\bar{N}^{\text{Total}}=235$; patches/$\text{WSI}$).
    $\bar{N}^{\text{Reduction}}$ is the mean number of patches retained after
    each method's selection step; the reduction percentage is computed
    relative to $\bar{N}^{\text{Total}}$.
}
\label{tab:patch_reduction_50pt}
\end{table}

\begin{table}[t]
\centering
\begin{tabular}{lccc}
\toprule
Setting & $\bar{N}^{\text{Total}}$ & $\bar{N}^{\text{Reduction}}$ & Reduction \\
\midrule
Yottixel+Pathologist  & 171 & 10 & 93.9\% \\
SPLICE+Pathologist    & 171 & 11 & 93.5\% \\
TITAN+Pathologist     & 171 & 171 & 0.0\% \\
PRISM+Pathologist     & 171 & 171 & 0.0\% \\
\midrule
CRISP ($s_t{=}30,\,K{=}7,\,\alpha{=}8.5\%$) & 525 & 7 & \textbf{98.6\%} \\
\bottomrule
\end{tabular}
\caption{Patch reduction statistics. For baselines (Yottixel, SPLICE, TITAN, PRISM),
each case contributes a single key-slide WSI;
$\bar{N}^{\text{Total}}=171.2$ patches/WSI.
For CRISP, all available WSIs per case are pooled from the full multi-WSI
dataset; $\bar{N}^{\text{Total}}=525$ patches/case (109,725 total across
641~WSIs, 209 patients).
$\bar{N}^{\text{Reduction}}$ is the average number of patches retained after
the method's selection step. Reduction is computed from $\bar{N}^{\text{Total}}$.}
\label{tab:patch_reduction_data2}
\end{table}

\section{Analysis and Discussions}

\textbf{Diagnosis -- }At the whole-slide image (WSI) level, TITAN and PRISM generally achieved slightly better performance for diagnostic subtyping. This suggests that methods that process a large number of patches and aggregate information across the entire slide can be effective for WSI-level diagnostic prediction. However, Yottixel achieved superior results in some settings, outperforming both TITAN and PRISM. This finding indicates that exhaustive patch-level processing followed by global aggregation may not always be optimal for diagnostic tasks at the WSI level. 
One possible explanation is that aggregating information from all patches may dilute diagnostically relevant signals, particularly when the features that define a subtype are focal, spatially limited, or present only in a small fraction of the tissue. In such cases, including a large number of non-informative or weakly informative patches may reduce the impact of the most relevant regions. By contrast, a method such as Yottixel's mosaic, which relies on only a subset of patches, may be better positioned to focus on the most discriminative areas of the slide. This could explain why Yottixel achieved improved performance in selected scenarios despite using less information overall. These results highlight the importance of patch-selection and attention mechanisms in WSI-level analysis and suggest that more data at the patch level does not necessarily translate into better diagnostic performance. 

Another important consideration is that TITAN, PRISM, Yottixel, and SPLICE are all WSI-level approaches that depend on the availability of a single representative WSI for each case. This requirement introduces practical and methodological challenges. First, selecting a representative slide can be time-consuming and may not always be feasible in routine clinical workflows, especially when multiple slides are available for a single patient or specimen. Second, the selection process may introduce subjectivity, because different pathologists may choose different slides as most representative of the case. This raises the possibility of inter-observer variability, which could influence model performance and limit reproducibility across institutions or users.

Therefore, while WSI-level models offer a practical framework for diagnostic prediction, their performance may depend not only on the model architecture but also on how slides and patches are selected and aggregated. Future studies should investigate the impact of representative-slide selection, inter-observer variability, and patch-sampling strategies on model performance. Evaluating these factors could help determine whether diagnostic models should process all available tissue, focus on selected high-yield regions, or incorporate mechanisms that adaptively identify diagnostically relevant patches across one or multiple WSIs.

\textbf{Treatment Outcome Prediction --} When considering treatment outcome, case-level processing with CRISP generally achieved better results than TITAN, PRISM, and Yottixel. This difference is notable because, unlike diagnostic subtyping, treatment outcome prediction is likely to depend on biological and morphological signals that may not be fully captured within a single WSI. TITAN, PRISM, and Yottixel are primarily WSI-level approaches; therefore, they rely on the analysis of one selected slide to represent the entire case. While this strategy may be sufficient or partially justified for diagnostic classification, it may be less appropriate for predicting treatment response or clinical outcome.

A likely explanation for the improved performance of CRISP is that it processes the case more comprehensively by incorporating information from all available WSIs. In doing so, CRISP may capture visual and biological clues that are distributed across different tissue sections. These clues may include regional heterogeneity, variable tumor microenvironmental patterns, differences in stromal or immune composition, necrosis, growth patterns, or other features that are not uniformly present across all slides. If only a single WSI is selected for analysis, such information may be missed, particularly when the predictive signal is focal or present in only a subset of the case material. This may explain why WSI-level methods that perform well for diagnosis do not necessarily achieve comparable performance for treatment outcome prediction.

The \underline{distinction} between \textbf{diagnostic subtyping} and \textbf{outcome prediction} is important. For diagnosis, the use of a single representative WSI may have some practical and conceptual justification, because morphological subtyping is routinely performed by pathologists and is often based on representative histologic sections. In that setting, a pathologist can select a slide that captures the dominant diagnostic morphology of the tumor. However, treatment outcome prediction is different. Response to therapy is not directly available to the pathologist through routine visual assessment in the same way that histologic subtype is. It may depend on complex and spatially heterogeneous biological features that are not obvious, not localized to the representative diagnostic slide, or not consistently recognized by human observers.

Therefore, if computational models are used to predict treatment outcome at the case level, it is natural that they \underline{should process all WSIs available for that case} rather than relying on a single selected slide. Comprehensive case-level analysis may allow the model to integrate information across the full histologic representation of the tumor and its surrounding tissue. This broader view may be particularly important for predictive tasks in which the relevant signal is weak, distributed, or biologically heterogeneous. The superior performance of CRISP supports this interpretation and suggests that case-level aggregation across multiple WSIs may be more suitable than WSI-level analysis for outcome prediction.

These findings also emphasize that the optimal computational strategy may depend on the clinical task. WSI-level models may be effective for diagnostic applications where a representative slide can reasonably capture the relevant morphology. In contrast, treatment outcome prediction may require a more comprehensive case-level framework that accounts for inter-slide heterogeneity and integrates information from the entire available tissue set. Future studies should further investigate how many slides are needed, which slides contribute most strongly to prediction, and whether adaptive slide-selection or case-level attention mechanisms can improve both performance and interpretability.

\textbf{Limitations --} All experiments were conducted on breast pathology slides. Whether the mosaic construction and retrieval pipeline generalizes to other organ sites, such as lung, prostate, or colorectal tissue, remains untested. However, CRISP is conceptually organ-agnostic and should, in principle, be applicable to a broad range of tissue image datasets. Nevertheless, the substantial morphological diversity across tissue types, as well as differences in whole-slide image (WSI) acquisition protocols, may require site-specific tuning of key parameters, including the SPLICE threshold and the number of $k$-means clusters. As an unsupervised method, CRISP offers a practical advantage in settings where annotated data are scarce. However, parameter calibration remains necessary to achieve optimal performance when adapting the framework to a new organ site or dataset.

A more comprehensive comparison with existing retrieval systems was also not feasible because, to our knowledge, no publicly available dataset provides the multi-WSI-per-case structure required for case-level retrieval. Existing public benchmarks are typically organized at the single-slide or patch level and therefore cannot be used to evaluate case-level or patient-level retrieval methods such as CRISP. This limitation reflects a broader challenge in data availability within computational pathology. Consequently, validation on independent case-level cohorts should be prioritized as such resources become available.

\section*{Methods}

\subsection*{Dataset}
All experiments were performed on the Mayo Clinic breast datasets. 
The diagnostic dataset was an institutional cohort of Formalin-Fixed Paraffin-Embedded~(FFPE) whole
slide images comprising 50~patients, 523~WSIs, and 7~diagnostic
categories: benign ($n{=}10$), ductal carcinoma in situ~(DCIS, $n{=}5$), 
invasive carcinoma of no special type~(NST, $n{=}10$), invasive lobular carcinoma~(ILC, $n{=}10$),
mucinous carcinoma~(MC, $n{=}5$), fibroadenoma~(FA, $n{=}5$), and phyllodes tumor~(PT, $n{=}5$).
Each case contributes multiple WSIs acquired from distinct tissue
blocks (mean ${\approx}10.5$ WSIs per case), yielding a total of
523~slides across 50~cases. Diagnostic labels are assigned at the
case level. Two data configurations were used. The
\emph{full-metadata} cohort contained all 523~WSIs and was used for
the proposed case-level framework. The single-slide pathologist cohort
comprised one pathologist-designated WSI per case (50~WSIs total)
and was reserved exclusively for single-slide baseline evaluation.

A treatment dataset was also used in this study has 212 patients with triple-negative breast cancer (TNBC) drawn 
from the Mayo Clinic institutional archive. Three cases were excluded owing to absent 
pathological response labels, yielding a final dataset of 209 patients. Of these, 120 
(57.4\%) achieved pathological complete response (pCR, label~1) following neoadjuvant 
chemotherapy, while the remaining 89 (42.6\%) did not (non-pCR, label~0). Whole-slide 
images (WSIs) were available from two sources: 92 patients had longitudinal case-level 
slides acquired across multiple clinical time-points, contributing a 
median of 5.8 WSIs per case (range 2--18, total 532 slides); the remaining 117 
patients were represented by a single diagnostic key-slide each, for a 
combined dataset of 649 WSIs across the full cohort (mean 3.1 WSIs per case). 
This mixture of single- and multi-slide cases reflects the heterogeneity routinely 
encountered in real-world clinical archives, where the volume of digitized material 
varies substantially by institution and era of collection.

\subsection*{Framework overview}
Each patient's WSIs are processed in three stages: tissue 
segmentation and patch extraction, CRISP and deep feature 
extraction followed by retrieval. The CRISP design is in two stages: a redundancy reduction scheme, here SPLICE, controls within-slide morphological coverage and 
compactness for all case WSIs, while $k$-means and subsequent sampling, here Yottixel's mosaic, controls cross-slide integration 
and case-level diversity.

\subsection*{Patch extraction and tissue filtering}

Each WSI was tiled into a regular, non-overlapping grid of 256\,$\times$\,256 pixel patches at 5\,$\times$ magnification. Patches with tissue occupancy below 70\% were discarded, excluding background, ink
artefacts, and tissue-sparse border regions.
The retained patches for each slide serve as input to the CRISP
distillation procedure.

\subsection*{CRISP Stages}
CRISP performs two stages to extract a small set of patches from all WSIs that represent the case.

\paragraph{Stage~1: Redundancy Reduction}
We reduce within-WSI patch redundancies by comparing patches of a WSI with each other and eliminate redundant ones.  

RGB channel statistics at low magnification reliably capture 
staining differences that reflect tissue composition, without 
the cost of high-resolution feature extraction. For each 
retained patch, the per-channel mean and standard deviation 
are computed over all normalized pixel intensities of the 
$256{\times}256$~pixel region at $5\times$~magnification, 
forming a 6-dimensional descriptor. The mean 
encodes dominant staining tone; the standard deviation 
captures intra-patch color variation, separating 
morphologically uniform regions from those containing mixed 
tissue content. Patch selection proceeds through an iterative scan of all tissue patches on a given slide. 
The next patch available in raster-scan order becomes the current
reference. Euclidean distances in descriptor space are computed 
to all other active patches, and those falling below the 
$s_t$-th percentile are discarded as color-redundant. The 
reference is added to a ``collage'' and the scan moves to the 
next surviving patch, repeating until the active pool is empty.
The exclusion boundary recomputes at each step from the 
distance distribution of the current reference, so the 
admission criterion tracks local color variation rather 
than applying a fixed cutoff.

Second, the collage length is self-determined: slides with high staining diversity retain more patches, while morphologically uniform slides are
summarized by fewer, without any pre-specified patch-count target.
The splice threshold $s_t \in [0, 100]$ is the sole hyperparameter:
higher values produce smaller and more selective collages, while lower
values retain a broader sample of the slide's color space. In this
work, $s_t$ was explored over $\{20, 21, \ldots, 40\}$.
The output for each case is the union of all per-WSI collages, which is then forwarded to Stage~3 for case-level consolidation.

To perform this stage and generate a collage for each WSI, we used SPLICE to provide a small set of patches for each WSI \cite{alsaafin2024splice}.

\paragraph{Stage~2: Clustering and Sampling}
We mixed all collages delivered by redundancy reduction and grouped them in similar patch classes via clustering. Subsequently, we sample a certain percentage of each cluster to extract the "case mosaic". Per-slide collages from all WSIs of the same case are merged 
into a single pool, capturing staining variation, tissue 
compartment diversity, and spatial heterogeneity that no 
individual slide encodes in full. The $k$-means partitions this 
pool into $k$ color-similar groups; from each cluster, the 
$\alpha$\% of patches nearest the centroid are retained, with 
a minimum of one per cluster. Centroid-proximal selection 
ensures the final set covers each color mode present across 
the case rather than sampling arbitrarily within clusters. 
$k$ and $\alpha$ are the two hyperparameters for stage 2, explored  jointly in the grid search.

As an ablation experiment, SPLICE replaced 
$k$-means with a second sequential distillation pass over the 
pooled case collages, isolating the contribution of 
diversity-seeking clustering over redundancy suppression alone.

For stage 2, we employ, Yottixel's mosaic to perform a two-factor clustering~\cite{kalra2020yottixel}.

\subsection*{Deep feature extraction and retrieval}
After case-level patch selection and assembling the ``case mosaic'', high-resolution image regions corresponding to the retained patches were read from the source WSI at 20x magnification and encoded with a foundation model. We used  Virchow2  to embed the patches in the case mosaic. The 1280-dimensional CLS token served as the patch embedding.  
Following Yottixel~\cite{kalra2020yottixel}, 
case-to-case distances (case mosaic to case mosaic) were calculated to establish the median of minimum 
Euclidean distances across pairs of case mosaics to measure similarity of patient biopsies: each query patch was matched to its closest 
counterpart in the archive patient, and the median of those 
minimum distances defined the case-level score. The closest 
archive case was returned as the Top-1 result.
For majority-vote Top-$k$~(MV@$k$), the 
diagnostic or treatment outcome labels of the $k$ nearest archive cases 
were collected and the most frequent label was 
assigned as the diagnosis or prediction; $k \in \{1, 3, 5, 7\}$ was 
evaluated throughout.

\subsection*{Baselines}

Four single-slide baselines were evaluated using the pathologist-curated cohort: 1) Yottixel$+$Pathologist applied the original Yottixel idea to
the pathologist-selected single WSI per case. It constructed a per-slide patch mosaic via 
two-stage $k$-means: color-histogram clustering into $K{=}9$ 
groups followed by spatial sub-sampling, retaining 5\% of patches 
per color group, 2) SPLICE$+$Pathologist applied the slide-level SPLICE 
to the same pathologist-selected WSI, 3) TITAN as a bimodal foundation model applied in default configuration on pathologist-selected WSI, and 4) PRISM as a vision foundation model applied in default configuration on pathologist-selected WSI.

CRISP used all cases WSIs for each case.

\subsection*{Retrieval evaluation}
Performance was measured under Leave-One-Patient-Out~(LOPO)
cross-validation in each dataset: each case served as the sole query against
the remaining cases, with all WSIs of the held-out patient
withdrawn simultaneously to prevent self-comparisons.
Embeddings were 1280-dimensional Virchow2 CLS tokens; distance
search. Retrieval quality is
reported as macro-averaged F1 at Top-1, majority-vote
Top-3~(MV@3), and majority-vote Top-5~(MV@5), with macro
averaging applied to give equal weight to each of the
7~diagnostic classes and 2 treatment outcomes.

\subsection*{Implementation details}
The clustering of $k$-means
 was used from \texttt{scikit-learn} with a fixed random seed of~724
for reproducibility. For deep embeddings, the foundation model  \emph{Virchow2} inference was performed on a single NVIDIA GPU
using the \texttt{timm} library with a SwiGLU-activated ViT-H/14
architecture and four register tokens. A grid search was conducted over $21 \times 14 \times 40 = 11{,}760$
combinations, all grid-search evaluations ran on
CPU only, completing in 37.3~minutes.

\medskip
\noindent\textbf{Data availability.}
The Mayo Clinic breast datasets are institutional cohorts not publicly available due to patient privacy constraints.

\noindent\textbf{Code availability.}
Code and configuration files for reproducing all experiments will be
made available on Kimia Lab's GitHub page: https://github.com/KimiaLabMayo.

\section*{Acknowledgments}

The authors gratefully acknowledge the \emph{F. Craig and Patricia Jilk Fund for Data Science, Predictive Modeling \& AI for Breast Cancer} for supporting this study.
The authors also acknowledge the \emph{Mayo Clinic Comprehensive Cancer Center}, Rochester, MN, USA, for its ongoing support.

\section*{Author contributions statement}

Z.R.A. implemented the CRISP framework, performed the experiments, and drafted the manuscript. W.U. and S.A. contributed to the datasets and drafting the manuscript. S.Y., J.C.B, and M.P.G. provided pathological and clinical guidance, K.R.K and H.R.T. supervised the project. H.R.T conceived the approach and extensively revised the manuscript. All authors reviewed the final version.

\bibliography{references}

\begin{thebibliography}{10}
\urlstyle{rm}
\expandafter\ifx\csname url\endcsname\relax
  \def\url#1{\texttt{#1}}\fi
\expandafter\ifx\csname urlprefix\endcsname\relax\def\urlprefix{URL }\fi
\expandafter\ifx\csname doiprefix\endcsname\relax\def\doiprefix{DOI: }\fi
\providecommand{\bibinfo}[2]{#2}
\providecommand{\eprint}[2][]{\url{#2}}

\bibitem{madabhushi2016digital}
\bibinfo{author}{Madabhushi, A.} \& \bibinfo{author}{Lee, G.}
\newblock \bibinfo{journal}{\bibinfo{title}{Image analysis and machine learning in digital pathology: Challenges and opportunities}}.
\newblock {\emph{\JournalTitle{Medical Image Analysis}}} \textbf{\bibinfo{volume}{33}}, \bibinfo{pages}{170--175}, \doiprefix\url{10.1016/j.media.2016.06.037} (\bibinfo{year}{2016}).

\bibitem{kalra2020yottixel}
\bibinfo{author}{Kalra, S.} \emph{et~al.}
\newblock \bibinfo{journal}{\bibinfo{title}{Yottixel -- an image search engine for large archives of histopathology whole slide images}}.
\newblock {\emph{\JournalTitle{Medical Image Analysis}}} \textbf{\bibinfo{volume}{65}}, \bibinfo{pages}{101757}, \doiprefix\url{10.1016/j.media.2020.101757} (\bibinfo{year}{2020}).

\bibitem{tizhoosh2024image}
\bibinfo{author}{Tizhoosh, H.~R.} \& \bibinfo{author}{Pantanowitz, L.}
\newblock \bibinfo{journal}{\bibinfo{title}{On image search in histopathology}}.
\newblock {\emph{\JournalTitle{Journal of Pathology Informatics}}} \textbf{\bibinfo{volume}{15}}, \bibinfo{pages}{100375} (\bibinfo{year}{2024}).

\bibitem{tizhoosh2021searching}
\bibinfo{author}{Tizhoosh, H.~R.} \emph{et~al.}
\newblock \bibinfo{journal}{\bibinfo{title}{Searching images for consensus: can ai remove observer variability in pathology?}}
\newblock {\emph{\JournalTitle{The American journal of pathology}}} \textbf{\bibinfo{volume}{191}}, \bibinfo{pages}{1702--1708} (\bibinfo{year}{2021}).

\bibitem{lahr2024analysis}
\bibinfo{author}{Lahr, I.} \emph{et~al.}
\newblock \bibinfo{journal}{\bibinfo{title}{Analysis and validation of image search engines in histopathology}}.
\newblock {\emph{\JournalTitle{IEEE Reviews in Biomedical Engineering}}} \textbf{\bibinfo{volume}{18}}, \bibinfo{pages}{350--367} (\bibinfo{year}{2024}).

\bibitem{barker2016automated}
\bibinfo{author}{Barker, J.}, \bibinfo{author}{Hoogi, A.}, \bibinfo{author}{Depeursinge, A.} \& \bibinfo{author}{Rubin, D.~L.}
\newblock \bibinfo{journal}{\bibinfo{title}{Automated classification of brain tumor type in whole-slide digital pathology images using local representative tiles}}.
\newblock {\emph{\JournalTitle{Medical Image Analysis}}} \textbf{\bibinfo{volume}{30}}, \bibinfo{pages}{60--71}, \doiprefix\url{10.1016/j.media.2015.12.002} (\bibinfo{year}{2016}).

\bibitem{zhang2015hashing}
\bibinfo{author}{Zhang, X.}, \bibinfo{author}{Liu, W.}, \bibinfo{author}{Dundar, M.}, \bibinfo{author}{Badve, S.} \& \bibinfo{author}{Zhang, S.}
\newblock \bibinfo{journal}{\bibinfo{title}{Towards large-scale histopathological image analysis: Hashing-based image retrieval}}.
\newblock {\emph{\JournalTitle{IEEE Transactions on Medical Imaging}}} \textbf{\bibinfo{volume}{34}}, \bibinfo{pages}{496--506}, \doiprefix\url{10.1109/TMI.2014.2363661} (\bibinfo{year}{2015}).

\bibitem{hegde2019smily}
\bibinfo{author}{Hegde, N.} \emph{et~al.}
\newblock \bibinfo{journal}{\bibinfo{title}{Similar image search for histopathology: Smily}}.
\newblock {\emph{\JournalTitle{npj Digital Medicine}}} \textbf{\bibinfo{volume}{2}}, \bibinfo{pages}{56}, \doiprefix\url{10.1038/s41746-019-0131-z} (\bibinfo{year}{2019}).

\bibitem{abmil}
\bibinfo{author}{Shao, D.} \emph{et~al.}
\newblock \bibinfo{title}{Do multiple instance learning models transfer?}
\newblock In \emph{\bibinfo{booktitle}{International conference on machine learning}} (\bibinfo{year}{2025}).

\bibitem{lu2021clam}
\bibinfo{author}{Lu, M.~Y.} \emph{et~al.}
\newblock \bibinfo{journal}{\bibinfo{title}{Data-efficient and weakly supervised computational pathology on whole-slide images}}.
\newblock {\emph{\JournalTitle{Nature Biomedical Engineering}}} \textbf{\bibinfo{volume}{5}}, \bibinfo{pages}{555--570}, \doiprefix\url{10.1038/s41551-020-00682-w} (\bibinfo{year}{2021}).

\bibitem{shao2021transmil}
\bibinfo{author}{Shao, Z.} \emph{et~al.}
\newblock \bibinfo{title}{Transmil: Transformer based correlated multiple instance learning for whole slide image classification}.
\newblock In \emph{\bibinfo{booktitle}{Advances in Neural Information Processing Systems}}, vol.~\bibinfo{volume}{34}, \bibinfo{pages}{2136--2147} (\bibinfo{year}{2021}).

\bibitem{campanella2019clinical}
\bibinfo{author}{Campanella, G.} \emph{et~al.}
\newblock \bibinfo{journal}{\bibinfo{title}{Clinical-grade computational pathology using weakly supervised deep learning on whole slide images}}.
\newblock {\emph{\JournalTitle{Nature Medicine}}} \textbf{\bibinfo{volume}{25}}, \bibinfo{pages}{1301--1309}, \doiprefix\url{10.1038/s41591-019-0508-1} (\bibinfo{year}{2019}).

\bibitem{chen2024uni}
\bibinfo{author}{Chen, R.~J.} \emph{et~al.}
\newblock \bibinfo{journal}{\bibinfo{title}{Towards a general-purpose foundation model for computational pathology}}.
\newblock {\emph{\JournalTitle{Nature Medicine}}} \textbf{\bibinfo{volume}{30}}, \bibinfo{pages}{850--862}, \doiprefix\url{10.1038/s41591-024-02857-3} (\bibinfo{year}{2024}).

\bibitem{lu2024conch}
\bibinfo{author}{Lu, M.~Y.} \emph{et~al.}
\newblock \bibinfo{journal}{\bibinfo{title}{A visual-language foundation model for computational pathology}}.
\newblock {\emph{\JournalTitle{Nature Medicine}}} \textbf{\bibinfo{volume}{30}}, \bibinfo{pages}{863--874}, \doiprefix\url{10.1038/s41591-024-02856-4} (\bibinfo{year}{2024}).

\bibitem{vorontsov2024virchow}
\bibinfo{author}{Vorontsov, E.} \emph{et~al.}
\newblock \bibinfo{journal}{\bibinfo{title}{A foundation model for clinical-grade computational pathology and rare cancers detection}}.
\newblock {\emph{\JournalTitle{Nature Medicine}}} \textbf{\bibinfo{volume}{30}}, \bibinfo{pages}{2924--2935}, \doiprefix\url{10.1038/s41591-024-03141-0} (\bibinfo{year}{2024}).

\bibitem{xu2024gigapath}
\bibinfo{author}{Xu, H.} \emph{et~al.}
\newblock \bibinfo{journal}{\bibinfo{title}{A whole-slide foundation model for digital pathology from real-world data}}.
\newblock {\emph{\JournalTitle{Nature}}} \textbf{\bibinfo{volume}{630}}, \bibinfo{pages}{181--188}, \doiprefix\url{10.1038/s41586-024-07441-w} (\bibinfo{year}{2024}).

\bibitem{wang2024chief}
\bibinfo{author}{Wang, X.} \emph{et~al.}
\newblock \bibinfo{journal}{\bibinfo{title}{A pathology foundation model for cancer diagnosis and prognosis prediction}}.
\newblock {\emph{\JournalTitle{Nature}}} \textbf{\bibinfo{volume}{634}}, \bibinfo{pages}{970--978}, \doiprefix\url{10.1038/s41586-024-07894-z} (\bibinfo{year}{2024}).

\bibitem{zimmermann2024virchow2}
\bibinfo{author}{Zimmermann, E.} \emph{et~al.}
\newblock \bibinfo{journal}{\bibinfo{title}{Virchow2: Scaling self-supervised mixed magnification models in pathology}}.
\newblock {\emph{\JournalTitle{arXiv preprint arXiv:2408.00738}}}  (\bibinfo{year}{2024}).
\newblock \eprint{2408.00738}.

\bibitem{ding2025titan}
\bibinfo{author}{Ding, T.} \emph{et~al.}
\newblock \bibinfo{journal}{\bibinfo{title}{A multimodal whole-slide foundation model for pathology}}.
\newblock {\emph{\JournalTitle{Nature Medicine}}} \doiprefix\url{10.1038/s41591-025-03982-3} (\bibinfo{year}{2025}).

\bibitem{moozy}
\bibinfo{author}{Kotp, Y.}, \bibinfo{author}{Trinh, V. Q.-H.}, \bibinfo{author}{Pal, C.} \& \bibinfo{author}{Hosseini, M.~S.}
\newblock \bibinfo{title}{Moozy: A patient-first foundation model for computational pathology} (\bibinfo{year}{2026}).
\newblock \eprint{2603.27048}.

\bibitem{alsaafin2024splice}
\bibinfo{author}{Alsaafin, A.} \emph{et~al.}
\newblock \bibinfo{journal}{\bibinfo{title}{Splice -- streamlining digital pathology image processing}}.
\newblock {\emph{\JournalTitle{arXiv preprint arXiv:2404.17704}}}  (\bibinfo{year}{2024}).
\newblock \eprint{2404.17704}.

\bibitem{jain2011adh}
\bibinfo{author}{Jain, R.~K.} \emph{et~al.}
\newblock \bibinfo{journal}{\bibinfo{title}{Atypical ductal hyperplasia: interobserver and intraobserver variability}}.
\newblock {\emph{\JournalTitle{Modern Pathology}}} \textbf{\bibinfo{volume}{24}}, \bibinfo{pages}{917--923}, \doiprefix\url{10.1038/modpathol.2011.66} (\bibinfo{year}{2011}).

\bibitem{allison2014variability}
\bibinfo{author}{Allison, K.~H.} \emph{et~al.}
\newblock \bibinfo{journal}{\bibinfo{title}{Understanding diagnostic variability in breast pathology: lessons learned from an expert consensus review panel}}.
\newblock {\emph{\JournalTitle{Histopathology}}} \textbf{\bibinfo{volume}{65}}, \bibinfo{pages}{240--251}, \doiprefix\url{10.1111/his.12387} (\bibinfo{year}{2014}).

\bibitem{shaikovski2024prism}
\bibinfo{author}{Shaikovski, G.} \emph{et~al.}
\newblock \bibinfo{journal}{\bibinfo{title}{Prism: A multi-modal generative foundation model for slide-level histopathology}}.
\newblock {\emph{\JournalTitle{arXiv preprint arXiv:2405.10254}}}  (\bibinfo{year}{2024}).

\end{thebibliography}

\newpage
\section{Appendix} 
\begin{figure}[htb]
    \centering
    
    \resizebox{0.3\textheight}{!}{
    \begin{minipage}{\linewidth}
        \centering
        
        \includegraphics[width=\linewidth]{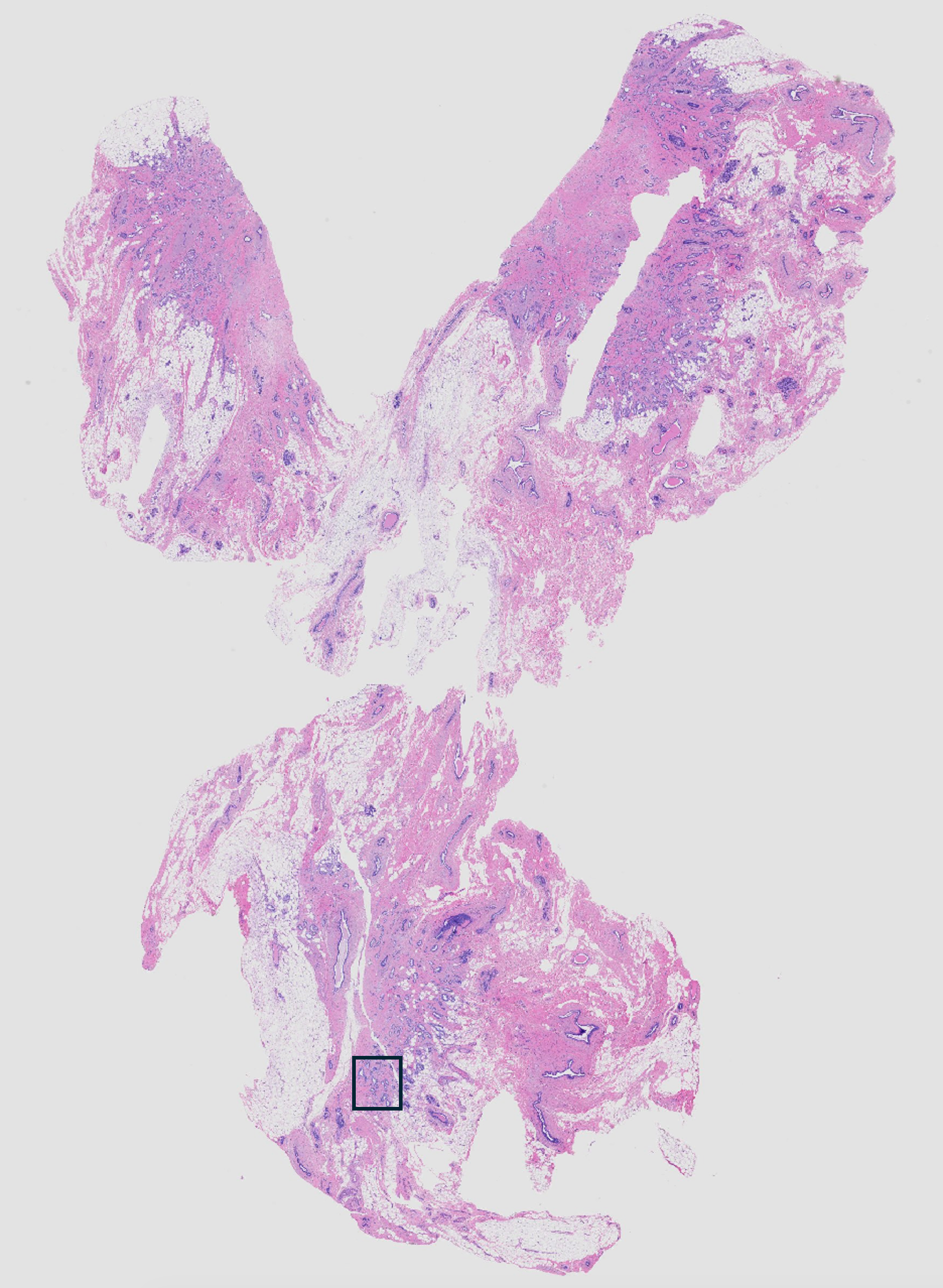}
        
        \vspace{0.5cm}
        
        \includegraphics[width=\linewidth]{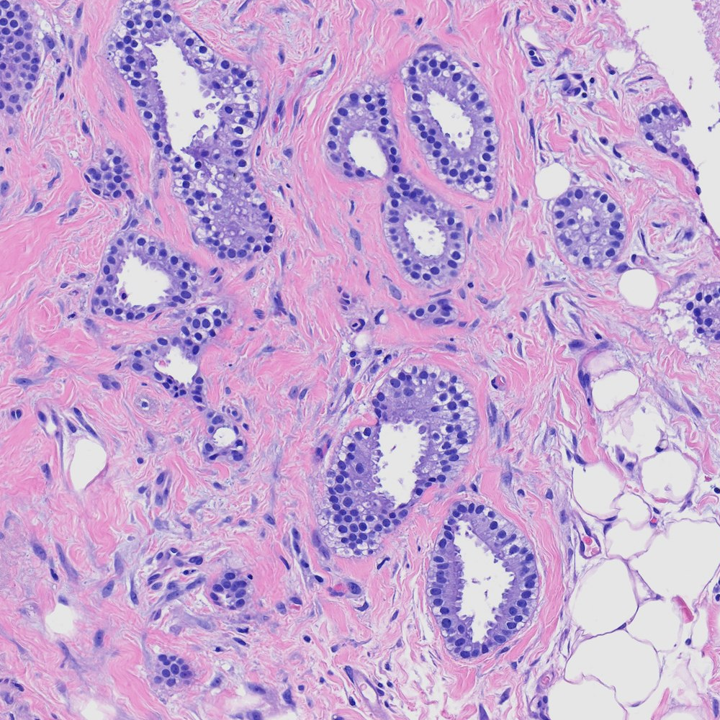}
    \end{minipage}
    }
    
    \caption{Top: Whole-slide image (WSI) for patient 1 from 50-patient dataset with the  subtype Invasive carcinoma, NST. Bottom: Corresponding patch extracted from the rectangular region selected in the top image at 20$\times$ magnification with a size of $1000 \times 1000$ pixels.}
    \label{fig:c01}
\end{figure}

\begin{figure}[htb]
    \centering
    
    \resizebox{0.6\textheight}{!}{
    \begin{minipage}{\linewidth}
        \centering
        
        \includegraphics[width=\linewidth]{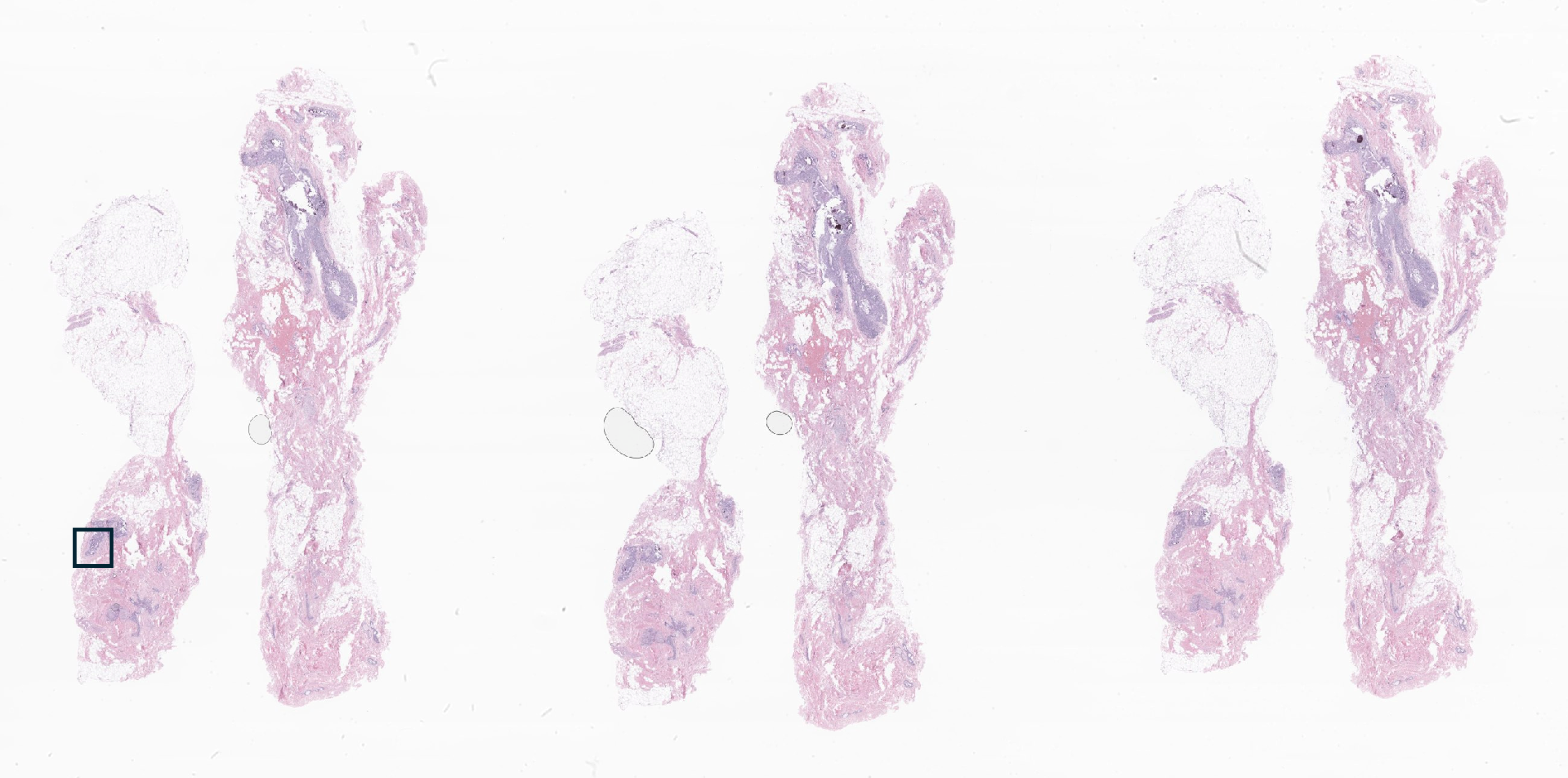}
        
        \vspace{0.5cm}
        
        \includegraphics[width=\linewidth]{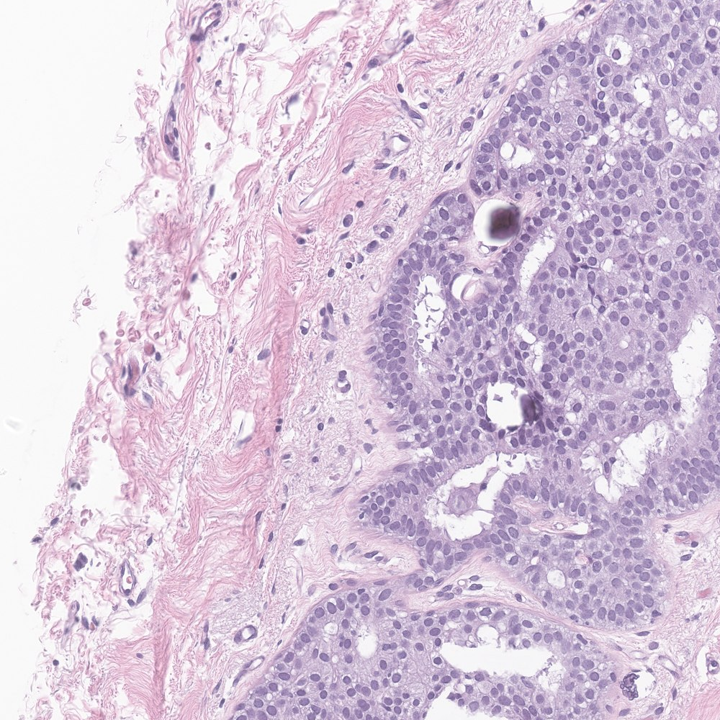}
    \end{minipage}
    }
    
    \caption{Top: Whole-slide image (WSI) for patient 33 from 50-patient dataset with the subtype Ductal carcinoma in situ. Bottom: Corresponding patch extracted from the rectangular region selected in the top image at 20$\times$ magnification with a size of $1000 \times 1000$ pixels.}
    \label{fig:c33}
\end{figure}

\begin{figure}[htb]
    \centering
    
    \resizebox{0.5\textheight}{!}{
    \begin{minipage}{\linewidth}
        \centering
        
        \includegraphics[width=\linewidth]{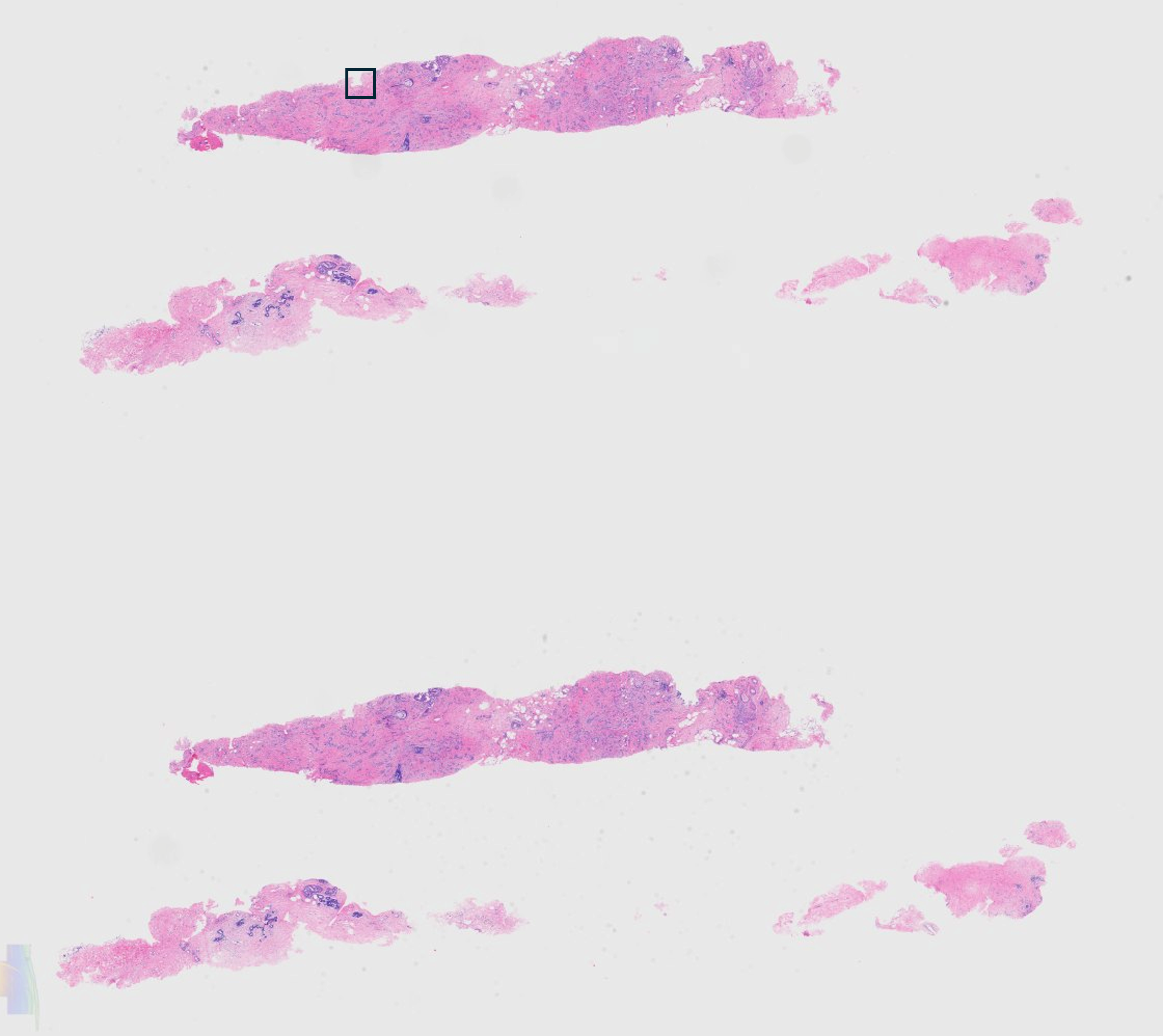}
        
        \vspace{0.5cm}
        
        \includegraphics[width=\linewidth]{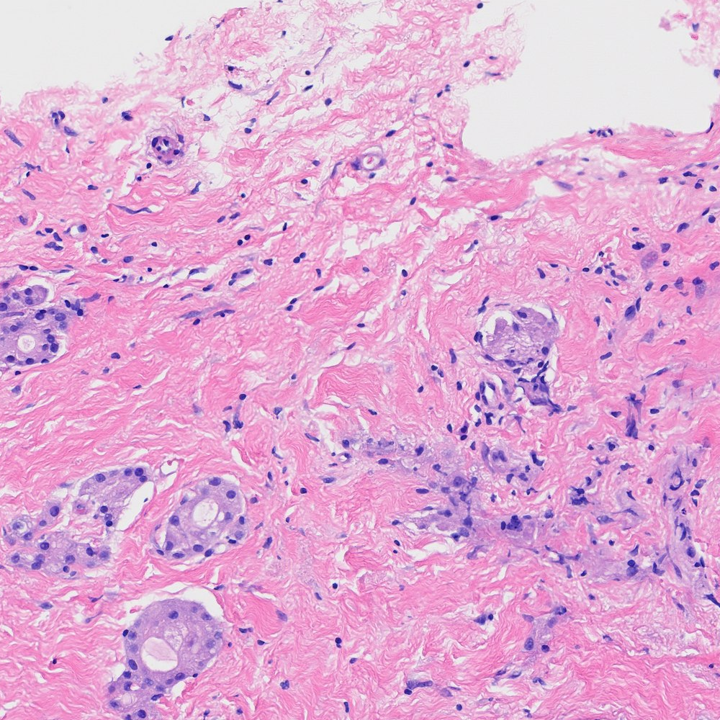}
    \end{minipage}
    }
    
    \caption{Top: Whole-slide image (WSI) for patient 1 from TNBC dataset with non-pCR. Bottom: Corresponding patch extracted from the rectangular region selected in the top image at 20$\times$ magnification with a size of $1000 \times 1000$ pixels.}
    \label{fig:tnbc1}
\end{figure}

\begin{figure}[htb]
    \centering
    
    \resizebox{0.5\textheight}{!}{
    \begin{minipage}{\linewidth}
        \centering
        
        \includegraphics[width=\linewidth]{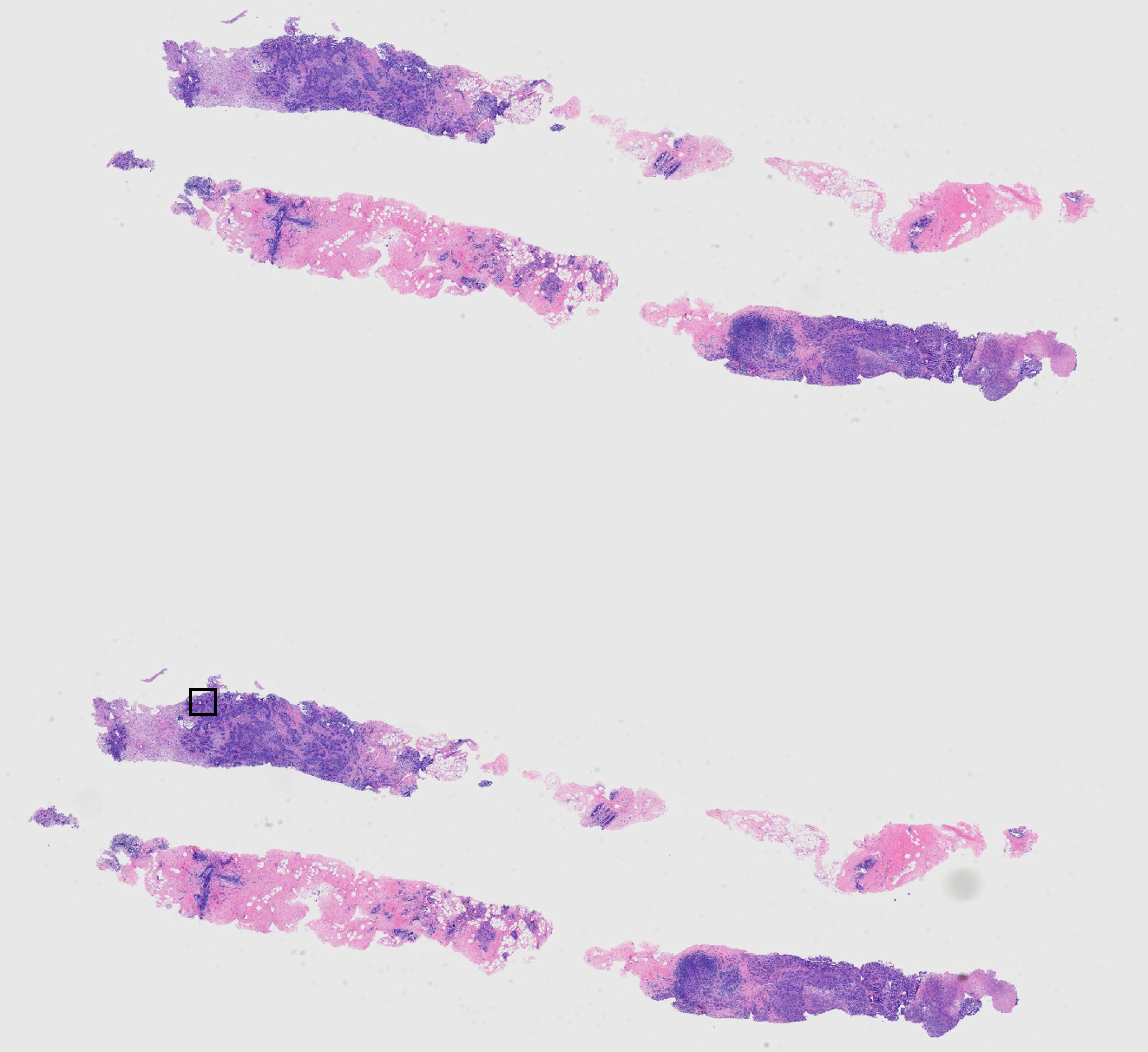}
        
        \vspace{0.5cm}
        
        \includegraphics[width=\linewidth]{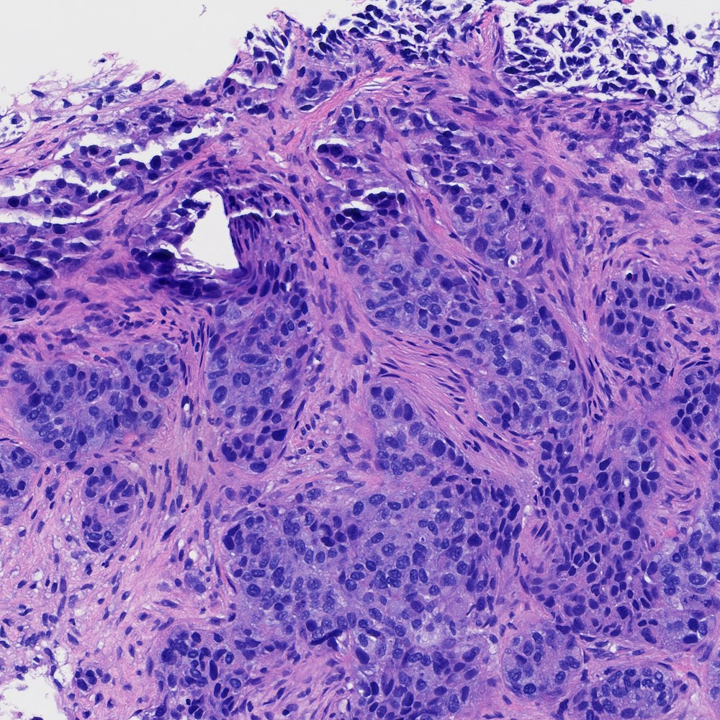}
    \end{minipage}
    }
    
    \caption{Top: Whole-slide image (WSI) for patient 21 from TNBC dataset with with pCR. Bottom: Corresponding patch extracted from the rectangular region selected in the top image at 20$\times$ magnification with a size of $1000 \times 1000$ pixels.}
    \label{fig:tnbc21}
\end{figure}

We present representative samples from the datasets used in this study. Figures~\ref{fig:c01} and~\ref{fig:c33} are from two different patients in the 50-patient dataset, corresponding to the subtypes Invasive carcinoma, NST and Ductal carcinoma in situ, respectively. Figures~\ref{fig:tnbc1} and~\ref{fig:tnbc21} are from the TNBC dataset and correspond to non-pCR and pCR cases, respectively.

\end{document}